\documentclass{article} % For LaTeX2e
\usepackage{iclr2026_conference}
\usepackage{times}      % 保留 Times 字体
\usepackage{amsmath,amsfonts,bm}

\def\eqref#1{equation~\ref{#1}}
\def\1{\bm{1}}

\DeclareMathAlphabet{\mathsfit}{\encodingdefault}{\sfdefault}{m}{sl}
\SetMathAlphabet{\mathsfit}{bold}{\encodingdefault}{\sfdefault}{bx}{n}

\usepackage{hyperref}
\usepackage{url}
\usepackage{booktabs}       % professional-quality tables
\usepackage{amsfonts}       % blackboard math symbols
\usepackage{siunitx}        % for S column type in tables
\usepackage{ulem}           % for \sout (strikeout text)
\usepackage{nicefrac}       % compact symbols for 1/2, etc.
\usepackage{microtype}      % microtypography
\usepackage{xcolor}         % colors

\usepackage{microtype}
\usepackage{graphicx}
\usepackage{amsthm}
\usepackage{booktabs}
\usepackage{algorithm}
\usepackage{algorithmic}

\usepackage{colortbl}
\usepackage{xcolor}
\usepackage{color}
\usepackage{footmisc}
\usepackage{enumitem}
\setlist{nolistsep}
\usepackage{xcolor}
\usepackage{diagbox}
\usepackage{makecell}
\usepackage{multirow}
\usepackage{booktabs}
\usepackage{listings}
\usepackage{makecell}
\usepackage{multicol}
\usepackage{wrapfig}
\usepackage[most]{tcolorbox} % 导言区引入
\usepackage{fancyvrb} 

\newtcolorbox{mybox}[1][]{%
  colback=gray!10,     % 背景色
  colframe=gray!80,    % 边框色
  coltitle=black,      % 标题颜色
  fonttitle=\bfseries, % 标题字体
  title=#1,            % 显示 title
  boxrule=0.8pt,       % 边框粗细
  arc=2mm,             % 圆角
  left=2mm, right=2mm, top=1mm, bottom=1mm % 内边距
}

\title{Elicit and Enhance: Advancing Multimodal Reasoning in Medical Scenarios}

\author{%
\textbf{Zhongzhen Huang}$^{*}$,
\textbf{Linjie Mu}$^{*}$,
\textbf{Yakun Zhu},
\textbf{Xiangyu Zhao},\\
\textbf{Shaoting Zhang}$^{\dag}$,
\textbf{Xiaofan Zhang}$^{\dag}$\\[1ex]
Shanghai Jiao Tong University
}

\iclrfinalcopy % Uncomment for camera-ready version, but NOT for submission.
\begin{document}

\renewcommand{\thefootnote}{\fnsymbol{footnote}}
\footnotetext{ \textsuperscript{*}Equal Contribution}
\footnotetext{ \textsuperscript{\dag}Corresponding authors}

\maketitle

\begin{abstract}
Effective clinical decision-making depends on iterative, multimodal reasoning across diverse sources of evidence. The recent emergence of multimodal reasoning models has significantly transformed the landscape of solving complex tasks. Although such models have achieved notable success in mathematics and science, their application to medical domains remains underexplored. In this work, we propose \textit{MedE$^2$}, a two-stage post-training pipeline that elicits and then enhances multimodal reasoning for medical domains. In Stage-I, we fine-tune models using a limited number of text-only data samples containing precisely orchestrated reasoning demonstrations to elicit reasoning behaviors. In Stage-II, we further enhance the model’s reasoning quality using rigorously curated multimodal medical cases, aligning model reasoning outputs with our proposed multimodal medical reasoning preference. Extensive experiments demonstrate the efficacy and reliability of \textit{MedE$^2$} in improving the reasoning performance of medical multimodal models. Notably, models trained with \textit{MedE$^2$} consistently outperform baselines across multiple medical multimodal benchmarks. Additional validation on larger models and under inference-time scaling further confirms the robustness and practical utility of our approach. 
\end{abstract}

% Medicine is a multifaceted endeavor involving daily diagnostic and therapeutic activities performed by healthcare professionals.

\section{Introduction}

Medicine is a multifaceted endeavor. It requires clinicians to process vast amounts of information, including patient medical histories, physical examination findings, and laboratory test results, to diagnose conditions, formulate prognoses, and determine appropriate treatment plans. In many clinical settings, an iterative reasoning process that evaluates multiple possibilities with progressively accumulated clinical information is considered fundamental to medical practice~\citep{adler2021next,singh2022five,croskerry2022advancing}. 
Recent advancements in multimodal large language models (MLLMs), such as OpenAI-o-series~\citep{openai2024reasoning} and Gemini-2.5-Pro~\citep{Google2025Gemini}, have significantly advanced complex task performance. These models employ scaling inference time and emulate reflective cognitive processes, pushing capabilities to unprecedented levels. In light of these advancements, we explore the question: \textit{How can we effectively extend the strategy of multimodal reasoning to medical domains}?

\begin{figure}[htb]
\centering
\scalebox{1}{
    \includegraphics[width=\linewidth]{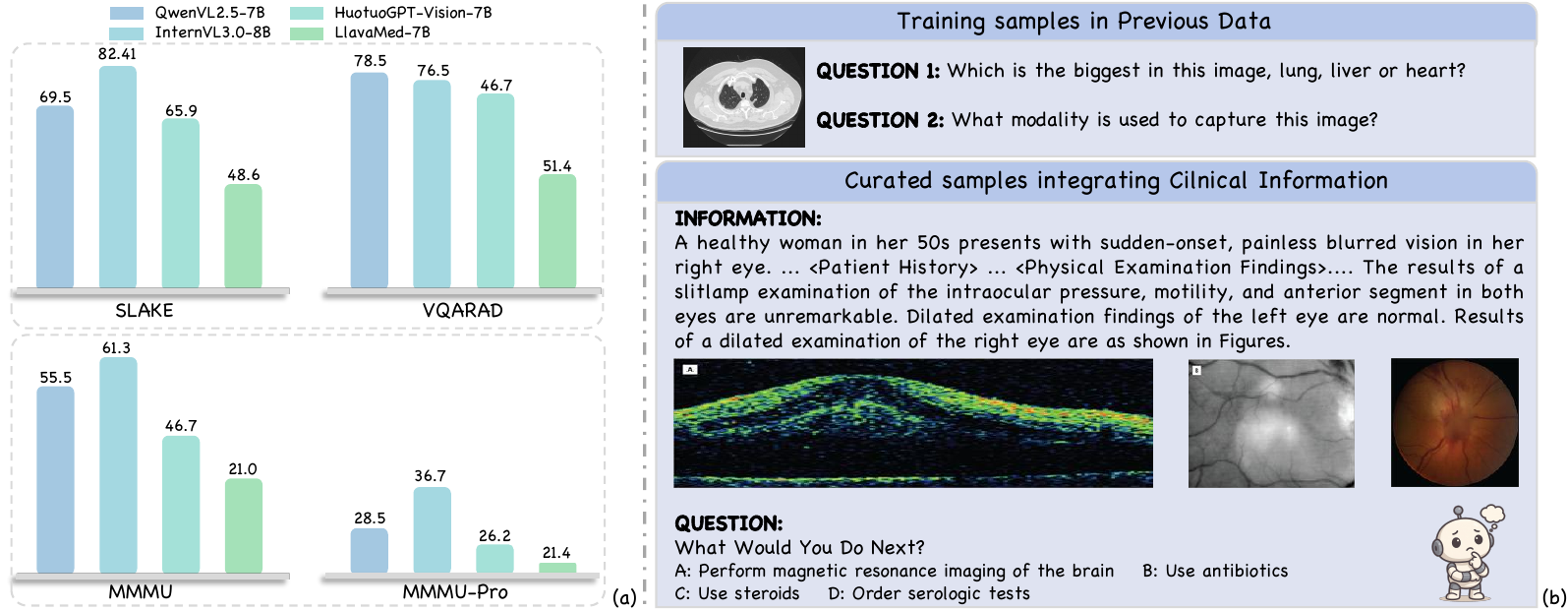}
}
\caption{
%General multimodal large language models are selected as the foundation for developing medical multimodal reasoning models because they consistently outperform models specifically tailored to medical domains. This performance gap widens as medical tasks become more diverse and challenging. To enhance multimodal reasoning capabilities, samples that closely reflect realistic clinical scenarios are prioritized, while those containing questions solvable primarily through pattern recognition or simple factual recall are excluded.
(a) Current models' performance on diverse tasks. (b) Samples that closely mirror real-world clinical scenarios are used to strengthen multimodal reasoning capabilities instead of samples focused primarily on pattern recognition or basic knowledge recall.
}
\vspace{-10pt}
\label{fig:intro}
\end{figure}

% Current multimodal reasoning models demonstrating remarkable proficiency are built upon powerful multimodal foundation models~\citep{bai2025qwen2,chen2024internvl} that perform well across various general benchmarks~\citep{yue2024mmmu,yue2024mmmupro,meng2025mm,he2024olympiadbench,lu2023mathvista}. 
% We begin by assessing the capabilities of different models within the medical domain. Such evaluations represent a natural starting point for probing a model’s medical knowledge and reasoning capacities. The results in left part of Figure~\ref{fig:intro} indicate that current multimodal models (both general~\citep{bai2025qwen2,chen2024internvl}  and specifically tailored to medical tasks~\citep{chen2024huatuogpt,li2023llava}) have demonstrate notable performance in simple visual question answering tasks~\citep{liu2021slake, lau2018dataset}. However, when it comes to complex tasks~\cite{yue2024mmmu, yue2024mmmupro} requiring more comprehensive understanding and reasoning capabilities, there is a notable performance degradation. Given that proficient reasoning models must be built on powerful foundation models~\citep{}, we adopt general MLLMs as a foundation for developing specialized medical reasoning models.

We begin by evaluating the capabilities of different models within the medical domain, as such assessments provide a natural starting point for probing a model’s medical knowledge and reasoning abilities. The results shown in the left panel of Figure~\ref{fig:intro} indicate that current multimodal models (i.e., general-purpose~\citep{bai2025qwen2,chen2024internvl} and those specifically designed for medical tasks~\citep{chen2024huatuogpt,li2023llava} ) demonstrate strong performance on relatively simple visual question answering tasks~\citep{liu2021slake,lau2018dataset}. However, performance declines markedly on more complex tasks~\citep{yue2024mmmu,yue2024mmmupro}, which require deeper comprehension and advanced reasoning. Since effective reasoning models should be grounded in robust foundation models~\citep{ye2025limo}, we adopt general-purpose MLLMs as the basis for developing specialized medical reasoning models.

% specifically tailored to medical tasks~\citep{chen2024huatuogpt,li2023llava} typically focus only on particular applications, such as visual question answering and medical report generation, and exhibit performance degradation in other medical tasks. 
% In contrast, general multimodal large language modelsdemonstrate superior results. Therefore, we adopt general MLLMs as a foundation for developing specialized medical reasoning models.

% To effectively enhance a model’s reasoning capabilities, the quality and relevance of training data are paramount, often outweighing the quantity of samples~\citep{deepseekai2025deepseekr1incentivizingreasoningcapability}. 

Previous research~\citep{deepseekai2025deepseekr1incentivizingreasoningcapability, ye2025limo, li2025limr} primarily relied on carefully curated, challenging datasets from mathematical and scientific domains to incentivize the model’s reasoning capacity. In contrast, although medical practice encompasses numerous scenarios requiring reasoning, the available data remain limited, especially those integrating clinical information. To alleviate this, we adopt a rigorous data construction process involving meticulous data collection, cleaning, and expert review. Unlike earlier studies~\citep{pan2025medvlm,lai2025med} that relied primarily on simple question–answer pairs solvable through memorized knowledge or pattern recognition (Figure~\ref{fig:intro}), our curated dataset comprises 3K textual questions and 2K multimodal questions spanning 12 imaging modalities, including radiology, pathology, and optical coherence tomography. These samples are sourced from authoritative examinations~\citep{jin2021disease} and publicly available case reports published in leading medical journals~\citep{pmc_open_access}. Each multimodal case integrates clinical histories, physical examinations, diagnostic investigations, and procedures.

With the dataset in place, we proceed to develop the models' reasoning capabilities specifically for clinical scenarios. Most of current work~\citep{meng2025mm,deepseekai2025deepseekr1incentivizingreasoningcapability,su2025gmai,li2025limr} usually employs rule-based reinforcement learning~\citep{shao2024deepseekmath} to advance models' reasoning capabilities. However, challenges such as normalization of specialized medical terms (e.g., anatomical terminology) and hierarchical relationships (e.g., organ-system classifications) in medical domains make it difficult to verify responses solely through predefined rules. Moreover, the prevalent strategy that relies on outcome-based rewards is vulnerable to hallucinations during the reasoning process. This poses significant risks in clinical applications, where diagnostic and therapeutic decision-making demands exceptional logical rigor and strong evidence-based justification. Building upon these, we introduce a novel training pipeline, named \textit{MedE$^2$}. Our pipeline involves two stages to progressively elicit and enhance the clinical cognitive reasoning capability of the models. Instead of \textit{cold-start} on multimodal data~\citep{deepseekai2025deepseekr1incentivizingreasoningcapability}, in the first stage, we conduct supervised fine-tuning on text-only data, where each sample includes carefully designed reasoning demonstrations to show how to utilize existing knowledge to solve complex tasks. Subsequently, we enhance the quality of the generated reasoning process to align with desired patterns. During this stage, we formulate Multimodal Medical Reasoning Preference (MMRP) and utilize Gemini-2.5-Pro to perform rejection sampling on our curated dataset, while employing Direct Preference Optimization (DPO)~\citep{ziegler2019fine, ouyang2022training} to calibrate and refine the model’s capabilities.

Extensive experiments are conducted to validate the effectiveness of our proposed pipeline. Despite being trained only on text-based reasoning data during Stage-I, the models demonstrate significant performance improvements across multiple medical benchmarks, achieving gains of at least 4.45\% on MedXpertQA-MM~\citep{zuo2025medxpertqa} and 6.67\% on MMMU-Pro-Health~\citep{yue2024mmmupro}. These results highlight an innovative method for eliciting reasoning behaviors in multimodal medical tasks. Further enhancement in Stage-II enables models to exhibit superior reasoning capabilities, reaching performance competitive with several larger-scale models. We also show that improvements from \textit{MedE$^2$} generalize well to models with larger parameters. Moreover, the consistent improvements observed under inference-time scaling further demonstrate the robustness of our post-training recipe. In summary, our main contributions are as follows:
% \vspace{10pt}
\begin{itemize}
    % \item We introduce \textit{MedE$^2$}, a reinforcement-learning-free, two-stage post-training pipeline aimed at advancing multimodal reasoning capabilities in medical scenarios. \textit{MedE$^2$} comprises two stages: (1) eliciting reasoning behavior using carefully orchestrated text-only data, and (2) enhancing reasoning quality with rigorously curated multimodal data.
    % \item We present \textit{MedE$^2$}, a two-stage post-training pipeline designed to enhance multimodal reasoning in medical scenarios. The pipeline comprises two critical stages: (1) eliciting reasoning behavior using strategically crafted text-only data, and (2) refining reasoning quality through meticulously curated multimodal data.
    \item We present \textit{MedE$^2$}, a two-stage post-training pipeline to enhance multimodal reasoning in medical scenarios: (1) eliciting reasoning behavior by leveraging strategically developed text-only datasets, and (2) refining the reasoning quality by incorporating meticulously curated multimodal data.
    \item We construct a high-quality dataset of approximately 5,000 samples, comprising both text-based questions with reasoning chains and multimodal questions involving clinical information, thereby establishing a reliable corpus for bootstrapping medical reasoning.
    \item We perform comprehensive experiments across diverse benchmarks. Experimental results consistently demonstrate that \textit{MedE$^2$} significantly enhances model performance and achieves strong generalization across models of varying sizes, while also being robust to inference-time scaling.
\end{itemize}

\section{Related Work}
\subsection{Medical Multimodal Large Language Models}
The development of multimodal large language models has significantly advanced the field of medicine. 
Current medical multimodal large language models are primarily based on general multimodal models and are further trained on specialized medical datasets. 
This approach led to the emergence of numerous medical multimodal large models. For example, Med-PaLM~\citep{tu2024towards} constructs the MultiMedBench dataset and fine-tunes based on PaLM-E~\citep{driess2023palm}. 
Recent studies continuously use similar strategies and refined training methods, resulting in significant progress, such as LLaVA-Med~\citep{li2023llava}, BioMedGPT~\citep{zhang2024generalist}, MedTrinity-25M~\citep{xie2024medtrinity}, and Med-Gemini~\citep{saab2024capabilities}. 
These strategies have demonstrated good application results in various medical scenarios, including medical dialogue~\citep{singhal2025toward}, clinical decision support~\citep{schubert2023evaluating}, electronic health record analysis~\citep{luo2022biogpt}, and image report generation~\citep{li2018hybrid}. As an endeavor that involves multi-level analysis of details, the reasoning ability is vital for solving medical tasks. In this study, we aim to explore how to incentivize the reasoning abilities of multimodal large models in medical tasks.

% Therefore, our study aims to explore how to enhance the reasoning abilities of multimodal large models in medical tasks.
% Despite these advancements, existing models still face challenges such as data scarcity and limitations in training methods. 
% The availability of high-quality multimodal medical data for training is limited, restricting the models' generalization capabilities. 

% Despite advancements, challenges such as limited high-quality multimodal medical data and insufficient training strategies for enhancing reasoning abilities hinder the full potential of these models.
% 多模态大模型的发展显著推动了医学领域模型的进步。当前的医学多模态大模型多基于通用大模型，通过专业医学数据集进行继续训练而构建。这一思路使得医学多模态大模型实现了大量涌现。例如，Med-Flamingo~\citep{moor2023med}在OpenFlamingo-9B基础上, 使用medical image-text data继续预训练。Med-PaLM~\citep{tu2024towards}则构建了MultiMedBench数据集，并基于PaLM-E进行微调。近年来，类似方法不断优化训练策略，取得了诸多进展，如LLaVA-Med~\citep{li2023llava}、BioMedGPT~\citep{zhang2024generalist}、MedTrinity-25M~\citep{xie2024medtrinity}、Qilin-Med-VL~\citep{liu2023qilin}和Med-Gemini~\citep{saab2024capabilities}等。这些模型已在多个医学场景中取得良好应用效果，, including scenarios such as medical dialogue~\citep{singhal2025toward}, clinical decision support~\citep{schubert2023evaluating}, electronic health record analysis~\citep{luo2022biogpt}, and image report generation~\citep{li2018hybrid}. 尽管取得了重大进展，现有医学多模态模型仍面临数据稀缺与训练方法受限的问题。可用于训练的高质量医学多模态数据有限，制约了模型泛化能力的提升。随着推理能力在科学任务中的成功应用，其在医学领域的潜力巨大，但目前尚缺乏有效的训练策略来增强医学模型的推理能力。因此，本研究旨在探索如何在有限医疗数据下，实现多模态大模型在医学任务中的推理泛化能力。

\subsection{Reasoning Models}

% This led to the introduction of heuristic search and the process-reward model. Heuristic search, inspired by algorithms such as Monte Carlo Tree Search~\citep{silver2017mastering}, extended single reasoning chains into tree-structured thoughts~\citep{yao2023tree, shinn2023reflexion}. In parallel, process-reward models framed reasoning as a Markov Decision Process, assigning stepwise rewards to guide the generation of more accurate reasoning chains~\citep{lightman2023let, jiao2024learning}.
Enhancing the reasoning abilities of models remains a key challenge. Early efforts evolved from few-shot prompting to structured paradigms like Chain-of-Thought~\citep{wei2022chain} and ReAct~\citep{yao2023react}, aiming to emulate human-like reasoning.
Subsequent work recognized reasoning as an iterative trial, error, and refinement process. The release of OpenAI's o1~\citep{openai2024reasoning} catalyzed further progress. Journey Learning~\citep{qin2024o1} explored strategies for o1-style slow thinking and STILL-2~\citep{min2024imitate} distilled long-form reasoning data to expand and refine solution paths. These advancements, culminating in 
DeepSeek's results~\citep{deepseekai2025deepseekr1incentivizingreasoningcapability}, highlight the critical role of reinforcement learning in improving reasoning. Diverse reinforcement learning (RL) approaches to further bolster reasoning are now being actively explored and investigated~\citep{wang2024enhancing, wei2025skywork}. However, effective training strategies to enhance the reasoning abilities of medical models are still lacking. Previous research has primarily focused on constructing reasoning processes or adapting reinforcement learning (RL) frameworks~\citep{su2025gmai, sun2025enhancingstepbystepverifiablemedical,pan2025medvlm,lai2025med}. In this paper, we propose a novel two-stage post-training recipe that progressively elevates models’ ability to perform fine-grained reasoning in the medical domain.

% LMM-R1~\citep{peng2025lmmr1}, Reason~\citep{tan2025reason}  Medical clinical scenarios inherently involve multi-level reasoning and analysis of details. With the continuous development of reasoning models, the application prospects of it are becoming increasingly evident. In this paper, we propose a noval two-stage post-training recipe that systematically elevates large models’ ability to perform fine-grained, multimodal reasoning in the medical domain.

\begin{figure}[tb]
\centering
\scalebox{1}{
    \includegraphics[width=\linewidth]{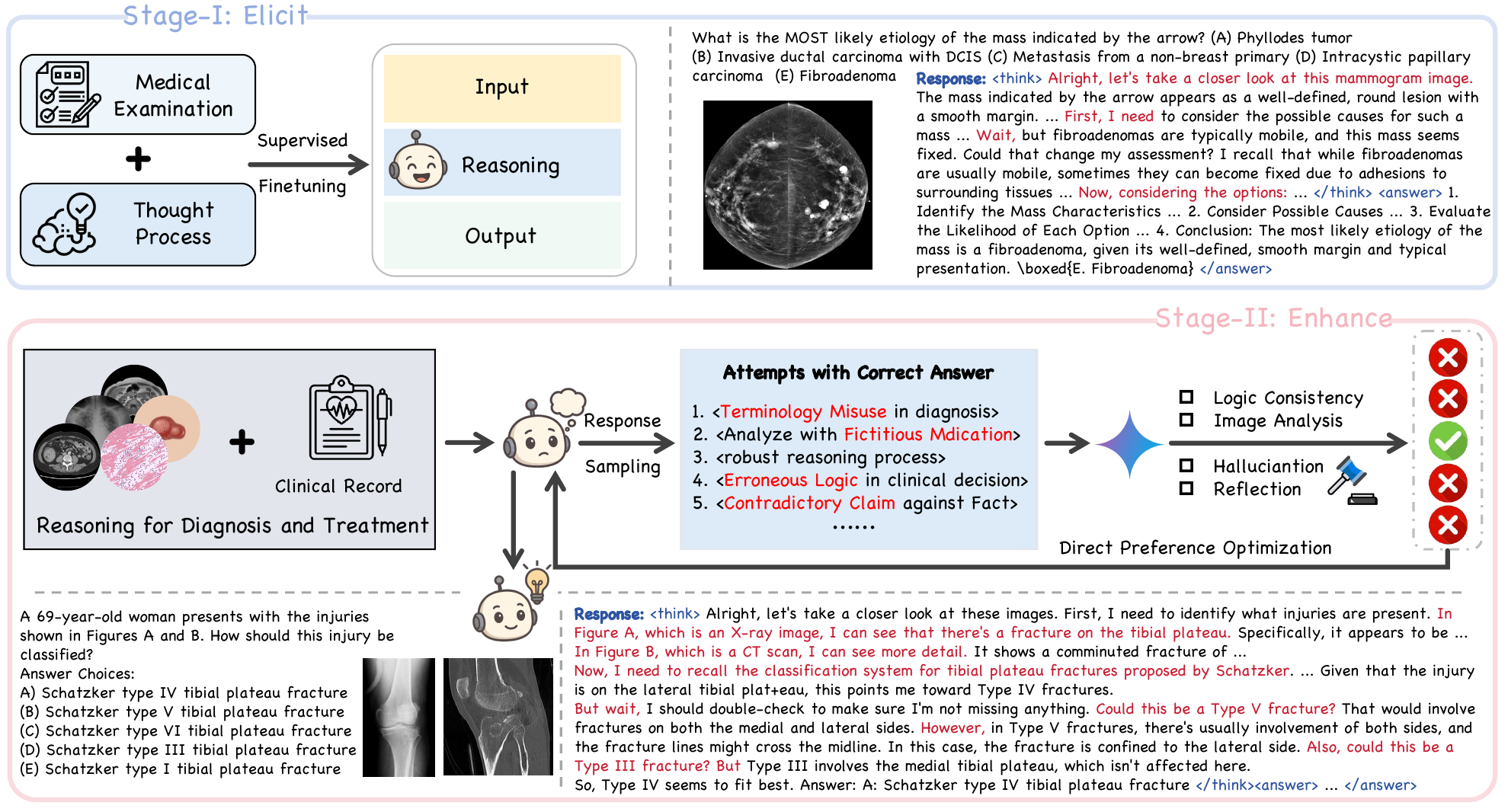}
}
\vspace{-10pt}
\caption{Overview of the two-stage post-training recipe \textit{MedE$^2$}. In Stage-I, text-only data containing reasoning demonstrations is employed to elicit initial reasoning behavior. In Stage-II, Direct Preference Optimization is applied to multimodal data to further enhance reasoning quality.}
\vspace{-10pt}
\label{fig:pipeline}
\end{figure}

\section{Pipeline}

This section presents the core pipeline of \textit{MedE$^2$}, as illustrated in Figure \ref{fig:pipeline}. To enhance the model's reasoning capabilities, we first curate a high-quality dataset comprising challenging text and multimodal question-answer pairs (Section~\ref{sec:data_collection}). Based on our pilot studies, we select general MLLMs (e.g., QwenVL-2.5~\citep{bai2025qwen2} and InternVL-3.0~\citep{chen2024internvl}) as the base models due to their performance across various medical tasks. Building upon these base models, we propose a two-stage training paradigm that progressively enhances the model's ability to reason in complex clinical scenarios (Section~\ref{sec:stage1} and Section~\ref{sec:stage2}).

\subsection{Data Collection}
\label{sec:data_collection}
We start with constructing a large-scale question bank sourced from authoritative examinations, top-tier journals, and publicly available datasets. In total, we collect 54K samples, consisting of 37K textual-only and 17K multimodal questions. To ensure high quality and prevent data leakage, a rigorous filtering process was implemented in collaboration with AI experts. 

Given that our goal is to explore the potential of multimodal reasoning strategies in solving complex medical tasks, we prioritize filtering out relatively simple samples. Specifically, we employ two baseline models (i.e., QwenVL-2.5 and InternVL-3.0) to eliminate samples that could be solved directly. For multimodal questions, we further prompt Gemini-2.5-Pro to identify the question type and exclude those classified as pattern-recognition tasks. To ensure that the remaining questions are solvable, AI experts independently attempted each question, with 4 attempts per model (Gemini-2.5-pro~\citep{Google2025Gemini} with DeepSeek-R1~\citep{deepseekai2025deepseekr1incentivizingreasoningcapability} for textual questions and Gemini-2.5-pro with Intern-S1~\citep{bai2025interns1scientificmultimodalfoundation} for multimodal questions). Only samples for which both experts answered correctly at least once were retained. Ultimately, this process resulted in a dataset consisting of 5K samples, including 3K textual and 2K multimodal questions.

% The remaining questions are further evaluated by Gemini-2.5-Pro, which attempts each question 4 times. We retain only those samples that the model answers correctly at least once. 

% Subsequently, two AI experts conduct independent sampling attempts on the remaining questions.  We preserved only those questions correctly answered by at least one expert across all attempts. Ultimately, this process resulted in a dataset consisting of 3.5K samples, including 2K textual and 1.5K multimodal questions.

\subsection{Stage-I: Eliciting Reasoning Ability}
\label{sec:stage1}
To enable models to effectively reason in clinical problem-solving scenarios, it is essential to instruct them on utilizing their existing knowledge base to address complex reasoning tasks. Previous studies~\citep{ye2025limo, li2025limr} have demonstrated that while supervised fine-tuning with reasoning-specific data can enhance reasoning task performance, reasoning capabilities can be elicited effectively with only a few illustrative examples. Indeed, since much of the relevant knowledge has already been encoded during pre-training, the construction of precisely orchestrated demonstrations of reasoning processes is more critical than merely increasing the volume of training data. In this stage, we utilize textual-only reasoning demonstrations to elicit sophisticated reasoning capabilities, rather than relying on multimodal reasoning data. This choice is motivated by empirical findings~\citep{su2025gmai} that entangled multimodal reasoning data can impair the model’s original language reasoning abilities. In contrast, training exclusively on textual data not only enhances reasoning skills but also maintains general visual understanding, albeit with a slight trade-off.

Building upon prior efforts~\citep{qin2024o1, huang2024o1, huang2025o1replicationjourney}, we adopt a distillation-based method to generate high-quality reasoning demonstrations using our curated dataset. Specifically, we leverage state-of-the-art reasoning models, including Gemini-2.5-pro, DeepSeek R1, and Qwen3-235B~\citep{qwen3technicalreport}, to produce diverse solutions. For open-source models, we directly use their generated reasoning processes. For the proprietary model, such as Gemini-2.5-pro, whose intermediate reasoning steps are not readily accessible, we further prompt gpt-OSS-120B~\citep{openai2025gptoss120bgptoss20bmodel} to expand its summarized outputs into complete reasoning processes. We conduct rigorous evaluations that combine rule-based filtering with human-assisted validation to ensure the quality of generated solutions. We design three criteria:  (i) the correctness of the final answer, (ii) logical structure and organization, and (iii) the plausibility and coherence of the reasoning process. Models trained on our curated textual reasoning data, denoted with the ``\texttt{Stage-I}'' suffix, demonstrate robust reasoning capabilities, not only in tackling textual domains but also when applied to multimodal scenarios.

% In addition, we select a subset of multimodal instances and prompt gpt-OSS-120B~\citep{} to elaborate the relevant explanations into detailed reasoning processes, which are filtered using the same criteria and used for comparison.

% Models trained on our curated textual reasoning data, denoted with the ``\texttt{Stage-I}'' suffix, demonstrate robust reasoning capabilities, not only in tackling textual domains but also when applied to multimodal scenarios.

\subsection{Stage-II: Enhancing Reasoning Quality}
\label{sec:stage2}
Although supervised fine-tuning can elicit reasoning behavior in models, it is often accompanied by an increase in hallucinations. This poses significant risks in medical scenarios, where clinical decision-making requires strong
evidence-based justification. Prior work~\citep{lv2024coarse,akbar2024hallumeasure} attributes this phenomenon to a mismatch between the generation paradigms during training and inference. Specifically, models are trained to predict the next token conditioned on preceding ground-truth tokens, while at inference time they rely on their own previously generated outputs. This discrepancy becomes particularly pronounced in tasks requiring long-form reasoning. As illustrated in Figure~\ref{fig:pipeline}, even a correct answer can emerge from a flawed reasoning process involving terminology misuse, logical errors, or fabricated medical content, etc. To align model outputs with human preference, the mainstream method is to employ Reinforcement Learning from Human Feedback (RLHF)~\citep{ziegler2019fine, ouyang2022training}, exemplified by Preference optimization (PO). In this work, we introduce Multimodal Medical Reasoning Preference (MMRP) and integrate it with Direct Preference Optimization (DPO)~\citep{rafailov2023direct, xu2024dpo}. This combination effectively reduces hallucinations and generates reasoning processes that align more closely with user requirements.

\textbf{Preliminary}
Considering a model as a policy $\pi_{\theta} (y|x)$ parameterized by $\theta$, RLHF aims at aligning the LLM $\pi_{\theta}$ with human preference. DPO is a representative algorithm of RLHF, serving as the preference
loss to optimize the policy $\pi_{\theta}$ by enabling the model to learn the relative preference between chosen and rejected responses. DPO eliminates the
requirement of training an explicit reward model based on
the assumption of the Bradley-Terry model~\citep{huang2004generalized} and directly optimizes $\pi_{\theta}$:
\[
\mathcal{L}_{\mathrm{DPO}}\left(\pi_{\theta}\right)=-\mathbb{E}_{\left(\mathbf{x}, \mathbf{y}_{w}, \mathbf{y}_{l}\right) \sim \mathcal{D}} 
\left[\log \sigma\left(\beta\log \frac{\pi_{\theta}\left(\mathbf{y}_{w} \mid \mathbf{x}\right)}{\pi_{\mathrm{ref}}\left(\mathbf{y}_{w} \mid \mathbf{x}\right)}-\beta\log \frac{\pi_{\theta}\left(\mathbf{y}_{l} \mid \mathbf{x}\right)}{\pi_{\mathrm{ref}}\left(\mathbf{y}_{l} \mid \mathbf{x}\right)}\right)\right]
\]
where $\mathbf{y}$ denotes the response of $\mathbf{x}$, and $\mathbf{y}_w$, $\mathbf{y}_l$ represent the ``win'' and ``lose'' items in preference pairs. $\sigma$ is the sigmoid function. $\pi_{ref}$ is the reference model used to regularize $\pi_{\theta}$ via Kullback–Leibler divergence, with $\beta$ controlling the strength of regularization.
DPO directly assigns higher probabilities to preferred responses, aligning with human preferences, while bypassing the train-inference mismatch present in SFT.

% \begin{tcolorbox}[colframe=black]
% \footnotesize\ttfamily
% - Logical Consistency: \\
% Is the reasoning process logically coherent and step-by-step valid?\\
% - Image Analysis Involvement: \\
% Does the reasoning process demonstrate appropriate analysis of any visual information mentioned or implied in the question?\\
% - No Hallucination: \\
% Does the reasoning process avoid introducing irrelevant or hallucinated information not supported by the question or known facts?\\
% - Reflection Presence: \\
% Does the reasoning show any self-checking, verification, or reflection on possible uncertainties?
% \end{tcolorbox}

\begin{tcolorbox}[
  colframe=gray!70, colback=gray!5,
  title=\textbf{Multimodal Medical Reasoning Preference}, % Box title
  fonttitle=\small\bfseries,
  coltitle=white,
  boxrule=0.8pt, arc=2mm, left=2mm, right=2mm,
  top=1mm, bottom=1mm,
  fontupper=\footnotesize\ttfamily % 字体设置
]

\begin{itemize}[leftmargin=*, nosep]
  \item The reasoning process is logically coherent and step-by-step valid.
  \item Contain appropriate analysis of relevant visual information.
  \item Avoid introducing hallucinated content not grounded in the input.
  \item Includes self-checking, verification, or reflection on uncertainties.
\end{itemize}

\end{tcolorbox}

\begin{wrapfigure}{r}{0.38\textwidth} % r=右侧 l=左侧
  \centering
  \includegraphics[width=0.35\textwidth]{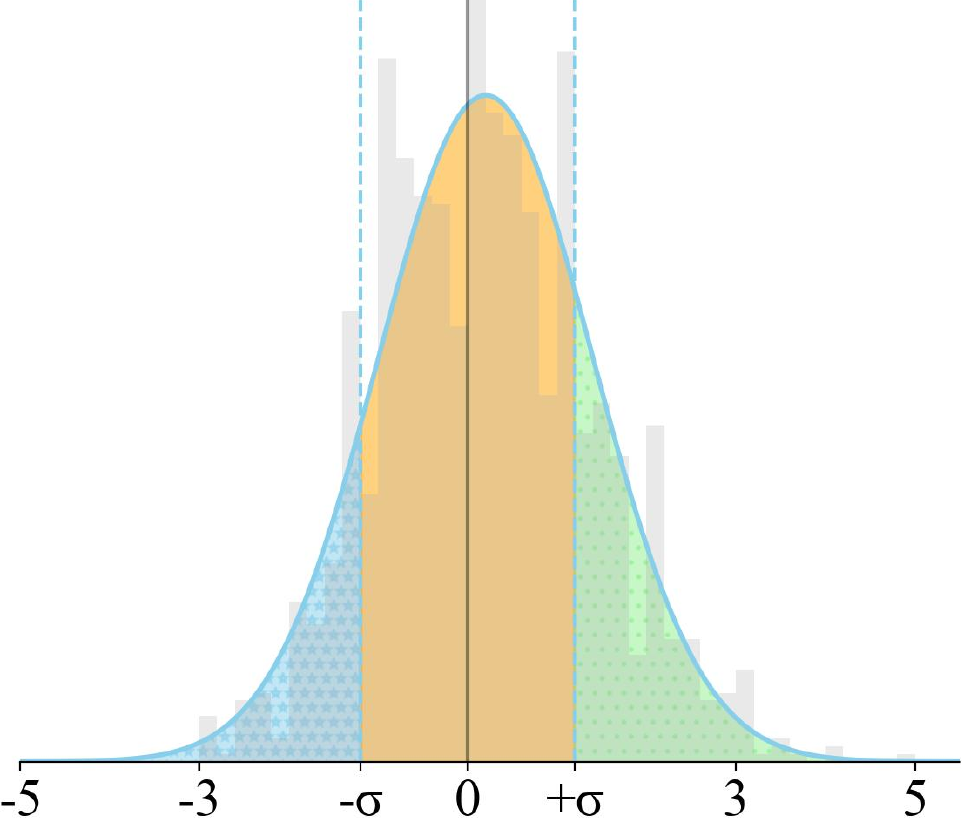}
  \vspace{-10pt}
  \caption{Distribution of human$–$model score differences, with 68.9\% falling within $\pm \sigma$, where $\sigma$=1.02.}
  \label{fig:dis}
\end{wrapfigure}

\textbf{Preference Data Construction}. Initially, we prompt the Stage-I model $M_1$ to generate multiple candidate reasoning processes for each sample in our multimodal training dataset. Given an image $I$ and a question $q$, we sample candidate reasoning processes $y$ from the distribution $M_1(y \mid q, I)$, repeating this process 8 times per sample to form a candidate set $Y$. Notably, each question $q$ undergoes meticulous refinement through collaboration between human experts and a model (GPT-4o), explicitly removing descriptions of image $I$ from $q$.  This ensures that reasoning arises from actual image examination rather than from textual captions. Subsequently, we filter out samples with wrong conclusions and employ Gemini-2.5-Pro to evaluate the remaining candidate reasoning processes according to four MMRP criteria: Logical Consistency, Image Analysis Involvement, Absence of Hallucinations, and Presence of Reflection. To validate the correctness of the model judgment, we select 1,000 cases and recruit human annotators to assess them using the same criteria. The differences between human scores and model judge scores are presented in Figure~\ref{fig:dis}. 
We find that model and human evaluations are largely consistent. Reasoning processes meeting all criteria form the positive set ($Y_p$), whereas those failing any criteria constitute the negative set ($Y_n$). We then construct preference pairs by selecting a preferred response $y_c$ from $Y_p$ and contrasting it with a response $y_r$ from $Y_n$. Unlike static dataset construction, our MMRP methodology enables dynamic and flexible curation of multimodal preference datasets, effectively balancing specialization and generalization in reasoning tasks. The prompt used for data evaluation with Gemini-2.5-Pro can be found in Appendix \ref{app:1}. Consequently, we obtain 4,432 pairs for QwenVL2.5-7B and 5,224 for InternVL3.0-8B. After applying DPO-based fine-tuning to the constructed preference dataset, the model demonstrates improved alignment with desired reasoning patterns and shows enhanced performance across diverse multimodal clinical scenarios.

\section{Experiments}

\subsection{Experimental Setup}

We select QwenVL2.5‑7B and InternVL3.0‑8B as the base models to evaluate the effectiveness of \textit{MedE$^2$}, resulting in two variants: (1) \texttt{+Stage-I}, which involves supervised fine-tuning on text-only data containing reasoning demonstrations; (2) \texttt{+Stage-II}, where the models from Stage-I are further enhanced using multimodal data and trained with DPO. We report the performance of GMAI‑VL‑R1~\citep{su2025gmai} and Chiron-o1~\citep{sun2025enhancingstepbystepverifiablemedical} as baseline methods for comparison. We also benchmark our approach against state-of-the-art open-source models that operate at substantially larger parameter scales \citep{bai2025qwen2, chen2024internvl} and several leading proprietary models—GPT-4o, OpenAI-o1, Gemini-2.5-Pro, and QvQ-Max~\citep{teamqvq}. Details of Implementation can be found in Appendix~\ref{app:exp}.

\textbf{Evaluation Benchmarks.} Our experiments involve two text-only tasks—MedQA~\citep{jin2021disease} and Medbullets~\citep{chen2025benchmarking}—as well as three multimodal benchmarks: MedXpertQA-MM~\citep{zuo2025medxpertqa}, MMMU-Health~\citep{yue2024mmmu}, and MMMU-Pro-Health~\citep{yue2024mmmupro}. Both MedQA and Medbullets are derived from questions used in the United States Medical Licensing Examination (USMLE). MedQA includes questions from both Step 1 and Step 2/3 of the Exam, while Medbullets focuses solely on Step 2/3, which are generally considered more challenging. For multimodal medical evaluation, MedXpertQA-MM serves as a highly challenging benchmark that assesses both medical knowledge and reasoning.

% All experiments are conducted on 8 NVIDIA H100 GPUs. LLaMA-Factory~\citep{zheng2024llamafactory} is used as the training framework, and vLLM~\citep{kwon2023efficient} serves as the backend during inference. 
% For all stages, models are fine-tuned using LoRA~\citep{hu2022lora} with a learning rate of $1 \times 10^{-4}$. Stage-II involves full-parameter fine-tuning with a learning rate of $1 \times 10^{-6}$, optimized using DeepSpeed~\citep{rasley2020deepspeed} with the ZeRO-3 configuration. Both stages adopt a cosine learning rate decay schedule.

% \subsection{Evaluation Benchmarks}
% Our experiments involve two text-only tasks—MedQA~\citep{jin2021disease} and Medbullets~\citep{chen2025benchmarking}—as well as three multimodal benchmarks: MedXpertQA-MM~\citep{zuo2025medxpertqa}, MMMU-Health~\citep{yue2024mmmu}, and MMMU-Pro-Health~\citep{yue2024mmmupro}. Both MedQA and Medbullets are derived from questions used in the United States Medical Licensing Examination (USMLE). MedQA includes questions from both Step 1 and Step 2/3 of the Exam, while Medbullets focuses solely on Step 2/3, which are generally considered more challenging. For multimodal medical evaluation, MedXpertQA-MM serves as a highly challenging benchmark that assesses both medical knowledge and reasoning. MMMU-Health and MMMU-Pro-Health are healthcare-specific subsets of MMMU and MMMU-Pro, respectively.

\begin{table}[t]
\centering
\label{tab:main}
\caption{Results (\%) of applying \textit{MedE$^2$} to \textcolor{orange!50}{QwenVL2.5-7B} and \textcolor{cyan!50}{InternVL3.0-8B} on multimodal medical benchmarks, along with state-of-the-art methods and other training strategies. $\Delta$ indicates the performance gain over the base model, and the best improvements are highlighted in bold.}
\resizebox{\linewidth}{!}{
    \begin{tabular}{l|cccc|cc|cc}
    \toprule
    \multirow{2}{*}{\textbf{Method}} 
    & \multicolumn{4}{c|}{\textbf{MedXpertQA-MM}} 
    & \multicolumn{2}{c|}{\textbf{MMMU-Health}} 
    & \multicolumn{2}{c}{\textbf{MMMU-Pro-Health}} \\
    \cmidrule(lr){2-5} \cmidrule(lr){6-7} \cmidrule(lr){8-9}
     & Reasoning & Understanding & Overall & $\Delta$
     & Overall & $\Delta$
     & Overall & $\Delta$ \\
    \midrule
    \multicolumn{9}{c}{\textit{\textbf{Proprietary}}} \\
    \midrule
    GPT-4o        & 40.73 & 48.19 & 42.80 & -- & 59.33 & -- & 40.91 & -- \\
    QvQ-Max       & 35.20 & 38.99 & 36.25 & -- & 70.67 & -- & 52.10 & -- \\
    OpenAI-o1        & 52.78 & 65.45 & 56.28 & -- & 60.00 & -- & 41.96 & --  \\
    Gemini-2.5-Pro & 61.69 & 69.13 & 63.75 & -- & 80.67 & -- & 66.08  & -- \\
    QwenVL2.5-32B & 26.00 & 30.86 & 27.35 & -- & 63.33 & -- & 48.25 & -- \\
    QwenVL2.5-72B & 27.10 & 31.58 & 28.35 & -- & 70.00 & -- & 50.35 & -- \\
    InternVL3.0-38B & 26.90 & 29.24 & 27.55 & -- & 68.00 & -- & 45.10 & -- \\
    InternVL3.0-78B & 28.80 & 35.92 & 30.20 & -- & 71.33 & -- & 51.40 & -- \\
    \midrule
    \multicolumn{9}{c}{\textit{\textbf{Baselines}}} \\
    \midrule
    \rowcolor{orange!8}QwenVL2.5-7B  & 19.99 & 22.56 & 20.70 & -- & 55.33 & -- & 28.47   & -- \\
    \rowcolor{cyan!8}InternVL3.0-8B  & 21.09 & 23.83 & 21.85 & -- & 61.33 & -- & 36.71 & --  \\
    \midrule
    \multicolumn{9}{c}{\textit{\textbf{Advancing}}} \\
    \midrule
    \rowcolor{orange!8}GMAI-VL-R1 SFT  &  -- & --  & 23.55 & + 2.85 & 56.00 & + 0.67 &  32.99   & + 4.52 \\
    \rowcolor{orange!8}GMAI-VL-R1 RLT &  -- &  -- & 23.80  & + 3.10 & 57.33 & + 2.00 & 34.03  & + 5.56 \\
    \rowcolor{orange!8}\texttt{+Stage-I} (\textbf{ours})  & 24.48 & 26.90 & 25.15 & + 4.45 & 62.00 & + 6.67 & 36.36   & + 7.89 \\
    \rowcolor{orange!8}\texttt{+Stage-II} (\textbf{ours}) & 25.80 & 28.52 & 26.55 & \textbf{+ 5.85} & 66.00 & \textbf{+ 10.67} & 38.81  & \textbf{+ 10.34} \\
    \midrule
    \rowcolor{cyan!8}Chiron-o1 & 23.30 & 25.10 & 24.20 & + 2.34 & 54.60 & - 6.72 & 33.90 & -2.80 \\
    \rowcolor{cyan!8}\texttt{+Stage-I} (\textbf{ours}) & 26.14 & 28.52 & 26.80 & + 4.95 & 69.33 & + 8.00 & 43.36 & + 6.65 \\
    \rowcolor{cyan!8}\texttt{+Stage-II} (\textbf{ours}) & 25.93     & 31.05     & 27.35     & \textbf{+ 5.50}  & 70.00    & \textbf{+ 8.67} & 48.95 & \textbf{+ 12.24} \\
    \bottomrule
    \end{tabular}
}
\end{table}

\subsection{Main Results}
\textbf{Text-only SFT Elicits Reasoning Behavior.} As shown in Table~\ref{tab:main}, even a limited number of samples formatted with extended reasoning chains can effectively elicit the reasoning behavior, resulting in substantial performance gains. For example, QwenVL2.5-7B achieves improvements of +4.45\% on MedXpertQA-MM and +6.67\% on MMMU-Health. More impressively, the performance improvements from reasoning elicitation are even more pronounced for stronger base models. For example, InternVL3.0-8B achieves an 8\% gain on MMMU-Health. This observation aligns with our hypothesis that training samples can serve as templates, illustrating how to utilize existing knowledge to solve complex reasoning tasks. Compared to \textsc{GMAI-VL-R1 SFT}, which relies on 10K multimodal samples, our method achieves a 6\% increase on MMMU-Health and a 3.37\% gain on MMMU-Pro-Health, using only half of the data in a text-only format. Similarly, relative to \textsc{Chiron-o1}, which relies on large-scale supervised fine-tuning with constructed CoT data, \textit{MedE$^2$} exhibits a clear performance advantage across all benchmarks. This underscores that high-quality, task-targeted supervision is more critical than sheer data quantity for developing reasoning abilities.

\textbf{Multimodal DPO Enhances Reasoning Quality.} When Direct Preference Optimization (DPO) is applied after SFT in Stage-II, it builds upon the structured reasoning patterns established during the earlier stage and further enhances the model's output quality. 
% Compared to reinforcement learning-based tuning (e.g., \textsc{GMAI-VL-R1 RLT}),  our method achieves notably better results: 26.55\% vs. 23.80\% on MedXpertQA-MM, 66.00\% vs. 57.33\% on MMMU-Health, and 38.81\% vs. 33.45\% on MMMU-Pro-Health. 
Compared to Group Relative Policy Optimization (GRPO)-based tuning~\citep{shao2024deepseekmath} (e.g., \textsc{GMAI-VL-R1 RLT}),  our method achieves notably better results: 26.55\% vs. 23.80\% on MedXpertQA-MM, 66.00\% vs. 57.33\% on MMMU-Health, and 38.81\% vs. 33.45\% on MMMU-Pro-Health.
A similar pattern is also observed in InternVL3.0-8B.
These results suggest that merely forcing the model to reason over limited-solution-space tasks during training is suboptimal, often leading to hallucinations or incoherent reasoning. Instead, activating reasoning capabilities requires not only complex tasks but also training samples that engage the model in inference-time computation, or in other words, examples with precisely orchestrated solutions. Although the performance gain from DPO is slightly smaller than those from text-only SFT, we observe that DPO helps regulate the issue of ``endless thinking'' and encourages the model to ``look before it thinks.''

\textbf{Comparison with State-of-the Art Models.} Table~\ref{tab:main} also presents a comparison with leading models. Among open-source models, those enhanced with \textit{MedE$^2$} demonstrate competitive or even superior performance despite having fewer parameters. For instance, on the MMMU-Health benchmark, QwenVL2.5-7B with \textit{MedE$^2$} achieves an accuracy of 66\%, outperforming its larger counterpart, QwenVL2.5-32B. The performance of InternVL3.0-8B with \textit{MedE$^2$} has already surpassed InternVL3.0-38B and is slightly lower than InternVL3.0-78B. Similarly, on the MedXpertQA-MM and MMMU-Pro-Health benchmarks, the smaller models with \textit{MedE$^2$} exhibit performance comparable to that of larger-scale models.

% 值得注意的是，InternVL3.0-8B with \textit{MedE$^2$}在
% Comparing the open-source modelsDepiste having fewer parameters, models with \textit{MedE$^2$}, demonstrates competitive or superior performance. For instance, 

\subsection{Ablation Studies}
\begin{figure}[tb]
\centering
\scalebox{1}{
\includegraphics[width=1\linewidth]{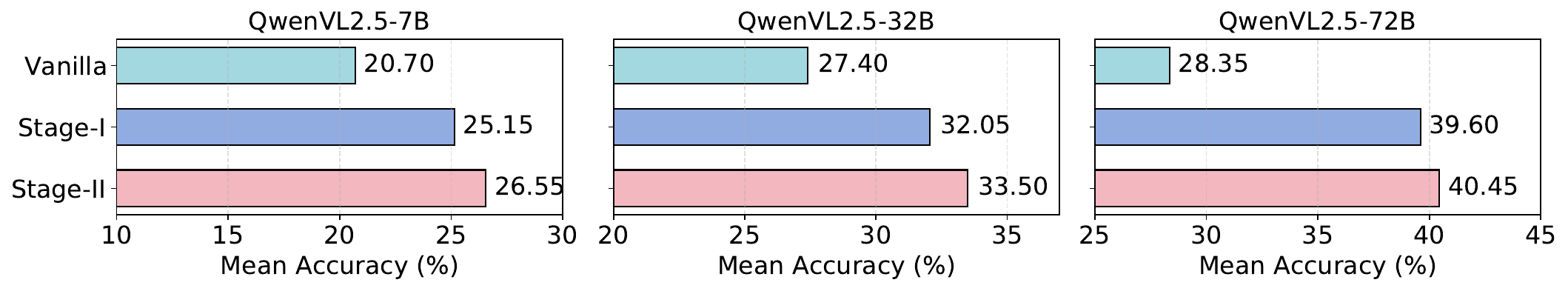}
}
\caption{The performance of QwenVL2.5-7B, 32B and 72B on the MedXpertQA-MM benchmark. As the model size increases, the models demonstrate progressively greater benefits from the \textit{MedE$^2$}.}
\label{fig: ab}
\vspace{-10pt}
\end{figure}

%Notably, these two larger models are fine-tuned using LoRA.
\textbf{Larger Models, Greater Improvements} Based on the preceding results, we observe that models exhibiting stronger initial performance tend to benefit more from reasoning elicitation. We hypothesize that larger models equipped with more extensive pretrained knowledge can derive increased benefits from our proposed \textit{MedE$^2$} framework. To rigorously verify this hypothesis, we apply our training pipeline to QwenVL2.5-32B and QwenVL2.5-72B and evaluate them on MedXpertQA-MM. We select MedXpertQA-MM for evaluation due to its increased challenge and minimal data leakage risk, as it was carefully constructed using difficulty-based filtering and data augmentation. 
Figure~\ref{fig: ab} illustrates the accuracy variations across different model sizes at various stages.
Following Stage-I, QwenVL2.5-72B demonstrates the most substantial performance gain, outperforming both QwenVL2.5-32B (4.65\%) and QwenVL2.5-7B (4.45\%). These findings are consistent with our assumption in Section~\ref{sec:stage1}: models with more extensive prerequisite knowledge encoded within their parameter space are more capable of learning reasoning patterns from limited yet precisely structured exemplars. Combining the results shown in Table~\ref{tab:main}, we surprisingly find that QwenVL2.5-72B enhanced with \textit{MedE$^2$} even surpasses QvQ-Max (40.45\% versus 36.25\%) on the MedXpertQA-MM benchmark, which is currently one of the most powerful models within the Qwen series. Although larger proprietary models such as o1 and Gemini-2.5-Pro still maintain an advantage on challenging benchmarks, the application of \textit{MedE$^2$} significantly narrows the performance gap between open-source and proprietary models. These results further demonstrate the effectiveness of \textit{MedE$^2$} in advancing multimodal reasoning capabilities for medical tasks. Detailed experimental results can be found in Appendix \ref{app:2}.

\textbf{Text-only vs. Multimodal Elicitation.} Previous studies have attempted to utilize multimodal training samples to activate multimodal reasoning capabilities. However, models optimized with such data often suffer from performance degradation on general vision tasks and impairment in linguistic abilities. 
% To better understand the differences between text-only and multimodal elicitation in the medical domain, we perform a direct comparison. 
% To better understand the distinctions between text-only and multimodal elicitation in the medical domain, we created a multimodal subset using the same methodology as in Stage-I. Subsequently, we prompted o1 to provide detailed reasoning processes as relevant explanations and conducted a direct comparison.
Specifically, we compare text-only SFT, multimodal SFT, and their combination within Stage-I. In addition to evaluating on multimodal benchmarks (MedXpertQA-MM and MMMU-Health), we also assess performance on two language-focused benchmarks: MedQA and Medbullets. As illustrated in Figure~\ref{fig:reduce_contamination}, both text-only and multimodal SFT successfully elicit reasoning behaviors and improve overall performance. However, multimodal SFT yields relatively smaller improvements compared to text-only SFT.  Notably, combining text-only and multimodal SFT results in reduced performance compared to the text-only SFT setting, particularly evident in the text tasks. This finding confirms that eliciting reasoning abilities through multimodal data may compromise the original language capabilities of the model.

% 调整两个子图空隙的
\setlength{\intextsep}{0pt}
\setlength{\floatsep}{0pt}

\begin{minipage}[htb]{0.47\linewidth} % 右侧，宽度为文本宽度的一半
    \begin{figure}[H]
        \centering
        \includegraphics[width=0.95\linewidth]{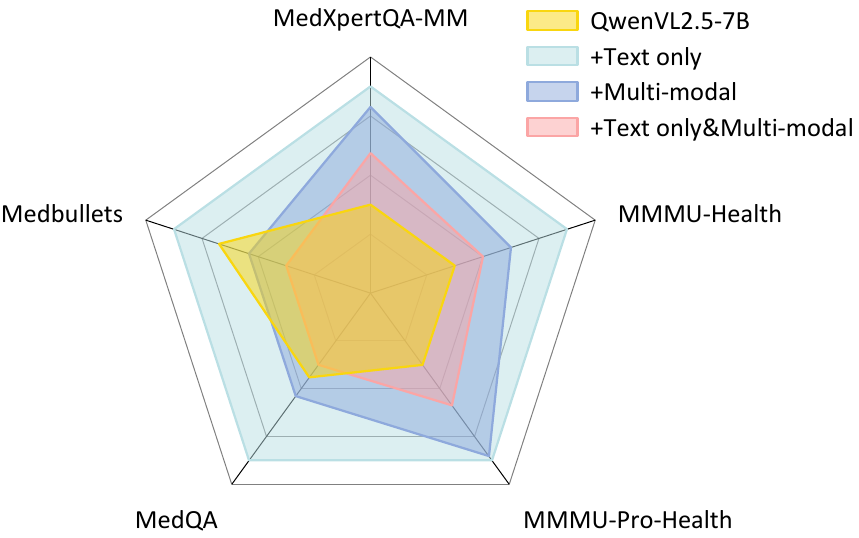} 
        % 增加点距离
        \setlength{\abovecaptionskip}{12pt} 
        \caption{Comparison of performance on Med-XpertQA-MM using various strategies for eliciting reasoning behaviors.}
        \label{fig:reduce_contamination}
        
    \end{figure}
\end{minipage}
\hspace{0.01\linewidth} % 控制两个 minipage 之间的水平间距
\begin{minipage}[tb]{0.5\linewidth} % 左侧，宽度为文本宽度的一半
    \begin{figure}[H]
        \centering
        \includegraphics[width=\linewidth]{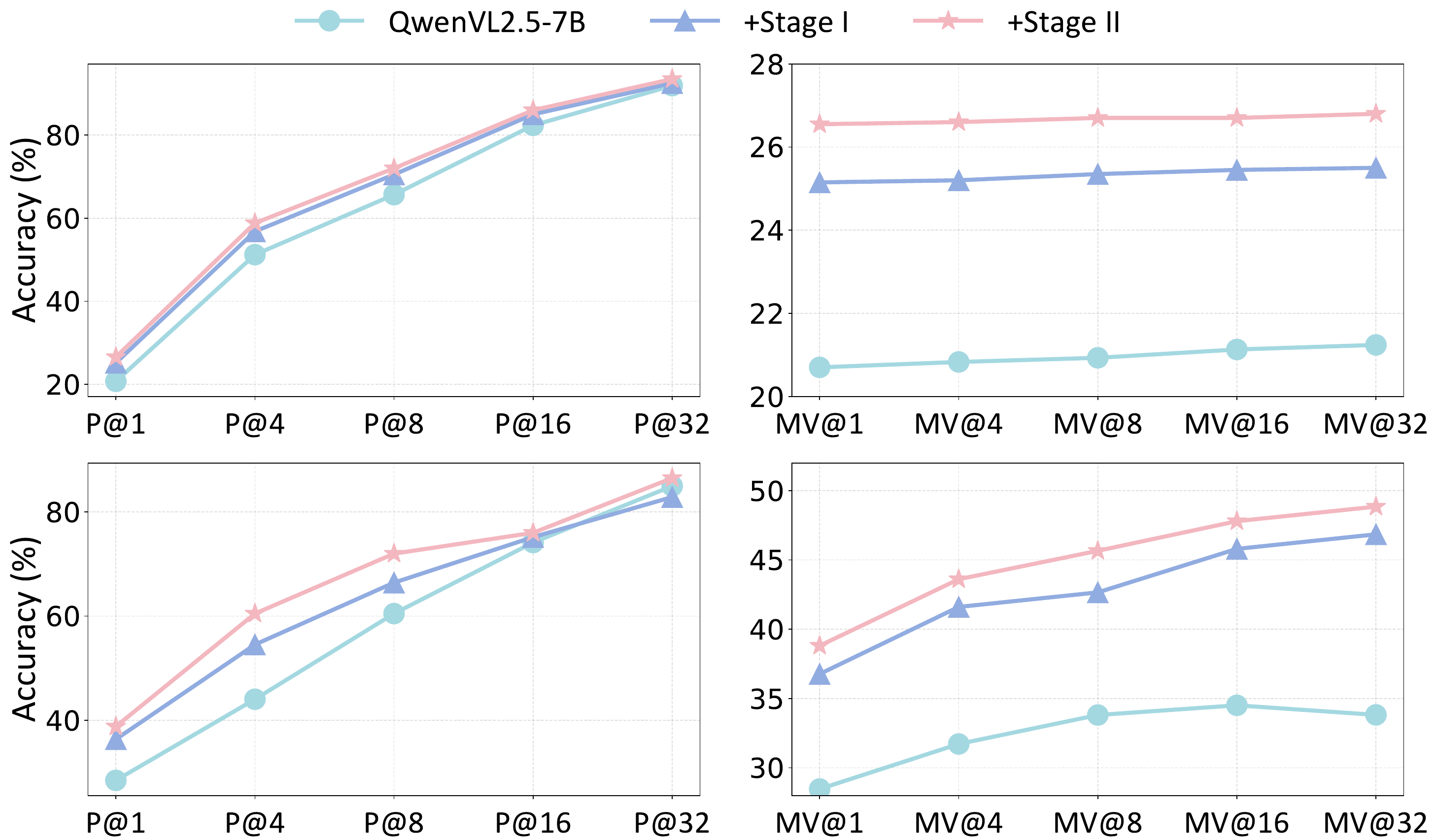} % 替换为你的PDF文件名
        \vspace{-10pt}
        \caption{The Accuracy of models with \textit{MedE$^2$} on MedXpertQA-MM (upper part) and MMMU-Pro-Health (lower part) with inference-time scaling.}
        \label{fig:radar_category}
    \end{figure}
\end{minipage}

\textbf{Benefit when Applying Inference Time Scaling.} To investigate the effectiveness of \textit{MedE$^2$}, in scaling up inference time, we report the results of majority-vote accuracy and pass@n accuracy on MedXpertQA-MM and MMMU-Pro-Health across 32 computational runs with temperature=0.5. The results, illustrated in Figure~\ref{fig:radar_category}, compare the performance of three models under Pass@n and majority voting (MV@n). Since both datasets are in a multiple-choice format with a limited answer space, model performance consistently improves as the number of attempts increases. Although the performance gain from majority voting on the MedXpertQA-MM dataset is modest due to its inherent difficulty, models trained with \textit{MedE$^2$} still demonstrate a significant and stable advantage as the number of computational attempts increases. On MMMU-Pro-Health, employing majority voting yields notable performance enhancements. Interestingly, the accuracy of QwenVL2.5-7B initially rises to a peak but then slightly declines as computational runs increase. In contrast, models trained with \textit{MedE$^2$} exhibit a steady performance improvement.  These findings suggest that \textit{MedE$^2$} enhances model output consistency, particularly evident during majority voting.
\begin{figure}[tb]
\centering
\scalebox{1}{
\includegraphics[width=\linewidth]{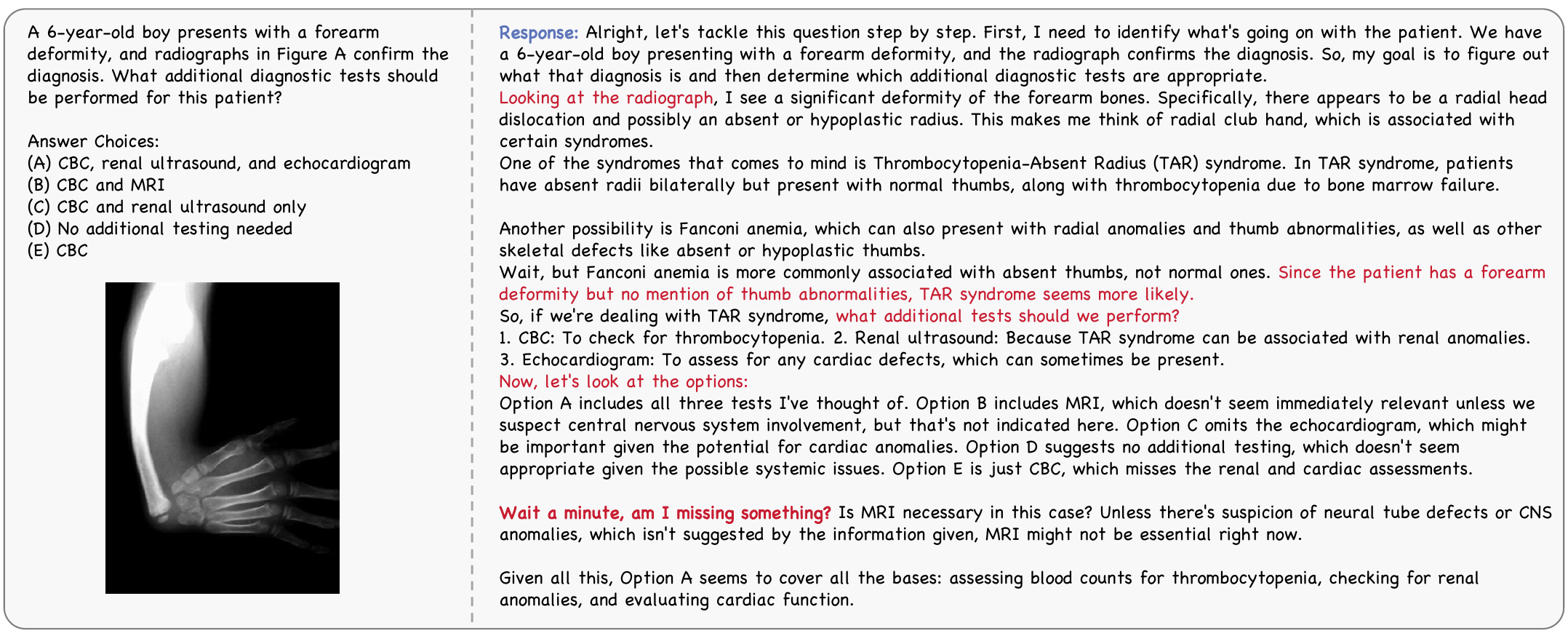}
}
\vspace{-10pt}
\caption{Case illustration of models' reasoning capability in tackling a clinical decision task.}
\vspace{-10pt}
\label{fig:example}
\end{figure}

\subsection{Qualitative Analysis}
Figure~\ref{fig:example} demonstrates the reasoning capabilities of Our\texttt{+Stage2} through its coherent analysis of determining appropriate follow-up examinations based on the patient's clinical presentation. In real-world scenarios, making an accurate diagnosis requires clinicians to integrate findings from multiple examinations to progressively narrow the differential diagnosis. Taking a closer look at the case, a notable observation is that the reasoning process is not constrained by the given options. Instead, it reflects a stepwise, hypothesis-driven diagnostic approach rather than merely discussing each option sequentially. Specifically, the model first identifies potential diagnoses (e.g., Thrombocytopenia-Absent Radius syndrome and Fanconi anemia) and then systematically excludes one by evaluating evidence from the radiographic findings. Decisions regarding subsequent examinations are logically derived from the provisional diagnosis, aligning closely with realistic clinical workflows. This observed capability highlights the potential of \textit{MedE$^2$} to serve as an effective post-training recipe to narrow the gap between current model performance and practical clinical application. Therefore, formulating an efficient and precise examination plan tailored to the patient's chief complaint is critical. We also remove the multiple-choice options and allow the model to respond freely, which can be seen in the Figure~\ref{fig:example_app2} in Appendix~\ref{app:3}.

\section{Conclusion}
In this study, we propose a novel reinforcement-learning-free training pipeline, \textit{MedE$^2$}, designed to progressively elicit and enhance multimodal reasoning capabilities in medical domains. \textit{MedE$^2$} comprises two distinct stages: it first performs supervised fine-tuning on carefully curated text-only medical reasoning data, followed by preference-based optimization using multimodal data through Direct Preference Optimization. To support medical reasoning training, we constructed a dataset of 5K questions spanning both textual and multimodal clinical scenarios. Extensive experiments demonstrate the effectiveness and reliability of \textit{MedE$^2$} in advancing the reasoning abilities of medical multimodal models. 
The demonstrated scalability and consistent improvements under inference-time scaling highlight its broad applicability across different model architectures and parameter scales. We hope that this early exploration into multimodal reasoning for medicine will inspire further research into specialized clinical reasoning capabilities.

\subsubsection*{Ethics Statement}
This work complies fully with the \href{https://iclr.cc/public/CodeOfEthics}{ICLR Code of Ethics}. All datasets used in this paper are publicly available and were obtained from established open sources. No private, sensitive, or personally identifiable information was collected or used. The study involves no human subjects, no experiments on vulnerable populations, and no interventions requiring IRB approval. We confirm that our methodology and results do not raise foreseeable risks of harm, misuse, or ethical concerns beyond standard scientific research practices.

\subsubsection*{Reproducibility Statement}
We will open-source the curated dataset and model weights at \url{https://anonymous.4open.science/r/MedEE-C381} to facilitate reproduction. We provide a comprehensive overview of our data construction methodology, the full processing pipeline, and the specific prompts used are provided in Appendix \ref{app:1}.
In addition, the training details and hyperparameter configurations for both stages of our method are presented in Appendix~\ref{app:exp}.

%%% END INSTRUCTIONS %%%

\bibliography{iclr2026_conference}
\bibliographystyle{iclr2026_conference}

\newpage
\appendix

\section{Prompt Templates and Data Formats Adopted in \textit{MedE$^{2}$}} 
\label{app:1}

\begin{tcolorbox}[
  colframe=gray!70,
  colback=gray!5,
  title=\textbf{The format of the text-only SFT data in Stage-I},
  fonttitle=\small\bfseries,
  coltitle=white,
  boxrule=0.8pt,
  arc=2mm,
  left=2mm,
  right=2mm,
  top=1mm,
  bottom=1mm
]
\begin{Verbatim}[fontsize=\footnotesize]
"messages": [
{
    "role": "system",
    "content": "You are a helpful assistant. A conversation between 
    User and Assistant. The user asks a question, and the Assistant 
    solves it. The assistant first thinks about the reasoning proc-
    ess in the mind and then provides the user with the answer. The 
    reasoning process and answer are enclosed within <think> </think> 
    and <answer> </answer> tags, respectively, i.e., <think> 
    reasoning process here </think> <answer> answer here </answer>."
},
{
    "role": "user",
    "content": "{{Patient Information}} {{Question}}"
}
]
\end{Verbatim}
\end{tcolorbox}

\begin{tcolorbox}[
  colframe=gray!70,
  colback=gray!5,
  title=\textbf{Prompt used by the Preference Dataset creation process},
  fonttitle=\small\bfseries,
  coltitle=white,
  boxrule=0.8pt,
  arc=2mm,
  left=2mm,
  right=2mm,
  top=1mm,
  bottom=1mm,
  fontupper=\footnotesize\ttfamily   % 直接用等宽字体
]
\begin{Verbatim}[fontsize=\footnotesize]
You are an expert reasoning evaluator.
I will provide you:
    * A question with images.
    * The groundtruth of the question
    * A model-generated answer, including its reasoning process.
Your task is to critically evaluate the reasoning based on 
the following five aspects:
    1. **Answer Correctness**: Is the final answer correct based 
    on the question?
    2. **Logical Consistency**: Is the reasoning process logically 
    coherent and step-by-step valid?
    3. **Image Analysis Involvement**: Does the reasoning process 
    demonstrate appropriate analysis of any visual information 
    mentioned or implied in the question?
    4. **No Hallucination**: Does the reasoning process avoid 
    introducing irrelevant or hallucinated information not supported 
    by the question or known facts?
    5. **Reflection Presence**: Does the reasoning show 
    any self-checking, verification, or reflection on
    possible uncertainties?
Please return the evaluation in the following JSON format:
    {
      "Answer_Correctness": "Yes/No",
      "Logical_Consistency": "Yes/No",
      "Image_Analysis_Involvement": "Yes/No",
      "No_Hallucination": "Yes/No",
      "Reflection_Presence": "Yes/No"
    }

<Question> \%s </Question> 
<Groundtruth> \%s </Groundtruth> 
<Answer> \%s </Answer>
\end{Verbatim}
\end{tcolorbox}
\newpage
\section{LLM Usage Statement}
We used LLMs to refine the writing, including checking grammar, rephrasing, and correcting typos. To ensure the writing quality, we further check and refine the generated text. In the experimental part of this study, LLMs were employed for two specific tasks: (1) Data preparation, as described in Section~\ref{sec:stage1}, and (2) Reasoning processes judgment, detailed in Section~\ref{sec:stage2}.

\section{Experimental Details}
\label{app:exp}
\textbf{Implementation Details.} All experiments are conducted on 8 NVIDIA H100 GPUs. LLaMA-Factory~\citep{zheng2024llamafactory} is used as the training framework, and vLLM~\citep{kwon2023efficient} serves as the backend during inference. For Stage-I, models are fine-tuned using LoRA~\citep{hu2022lora} with a learning rate of $1 \times 10^{-4}$. Stage-II involves full-parameter fine-tuning with a learning rate of $1 \times 10^{-6}$, optimized using DeepSpeed~\citep{rasley2020deepspeed} with the ZeRO-3 configuration. Both stages adopt a cosine learning rate decay schedule. For models with larger parameters(i.e., 32B and 72B), we also employed LoRA in Stage-II.

\section{Experimental Results with Different Model Scales on MedXpertQA-MM}
\label{app:2}

Our method demonstrates more significant performance on larger-scale models, as shown in Table \ref{tab:2}. These findings are consistent with our observations in Section~\ref{sec:stage1}: models with more extensive prerequisite knowledge encoded within their parameter space are more capable of learning reasoning patterns from limited yet precisely structured exemplars.

\begin{table}[t]
\vspace{10pt}
\caption{Performance comparison of QwenVL2.5-32B and QwenVL2.5-72B with \textit{MedE$^{2}$} across different task categories on the MedXpertQA-MM benchmark. As shown in the figure, the accuracy of each subtask improves, with the larger 72B model exhibiting a more pronounced increase compared to the 32B model.}
\vspace{10pt}
\centering
\resizebox{\linewidth}{!}{
\begin{tabular}{l|ccc|cc|c|c}
\toprule
\textbf{Method} & \textbf{Treatment} & \textbf{Basic Science} & \textbf{Diagnosis} & \textbf{Reasoning} & \textbf{Understanding} & \textbf{Overall} & $\Delta$ \\
\midrule
QwenVL2.5-32B                       & 28.13 & 25.78 & 27.61 & 26.07 & 30.87 & 27.40 & +0.00 \\
\quad \texttt{+Stage-I} (\textbf{ours})     & 33.03 & 30.31 &  32.19   & 31.60 & 33.21   & 32.05 & +4.65 \\ 
\quad \texttt{+Stage-II} (\textbf{ours})    & 35.93 & 32.57 & 32.86 & 33.54    & 33.39    & 33.50    & +6.10    \\
\midrule
QwenVL2.5-72B                       & 30.36 & 24.08 & 28.86 & 27.11 & 31.59 & 28.35 & +0.00  \\
\quad \texttt{+Stage-I} (\textbf{ours})     & 40.40 & 41.93 & 38.62 & 39.21 & 40.61 & 39.60 & +11.25 \\
\quad \texttt{+Stage-II} (\textbf{ours})    & 41.52 & 42.49 & 39.45 & 39.63 & 42.60  & 40.45 & +12.10 \\
\bottomrule
\end{tabular}
}
\label{tab:2}
\end{table}

\newpage
\section{Case Study}
\label{app:3}
\begin{figure}[htb]
\centering
\scalebox{1}{
\includegraphics[width=\linewidth]{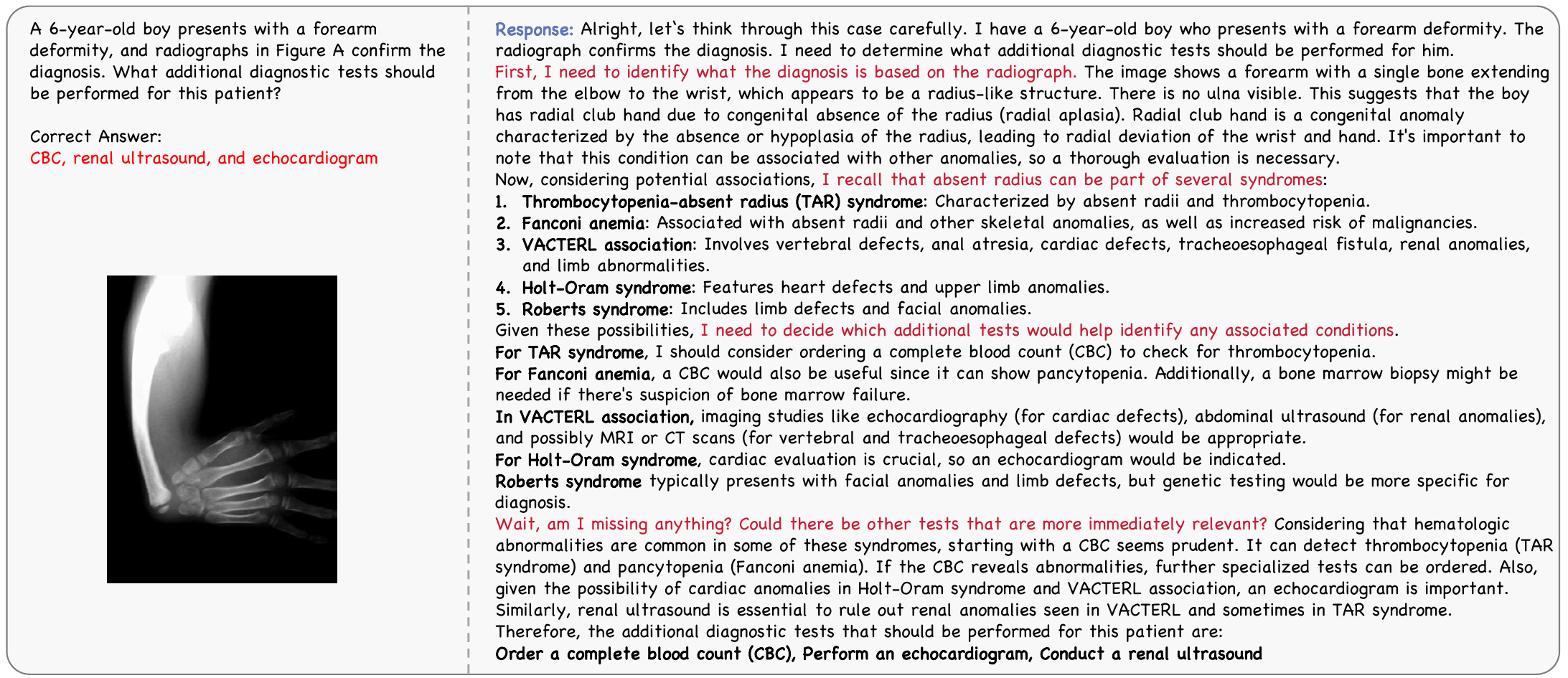}
}
\vspace{-10pt}
\caption{Case illustration of models' reasoning capability in the open-ended task.}
\label{fig:example_app2}
\end{figure}

\vspace{20pt}

\begin{figure}[hb]
\centering
\scalebox{1}{
\includegraphics[width=\linewidth]{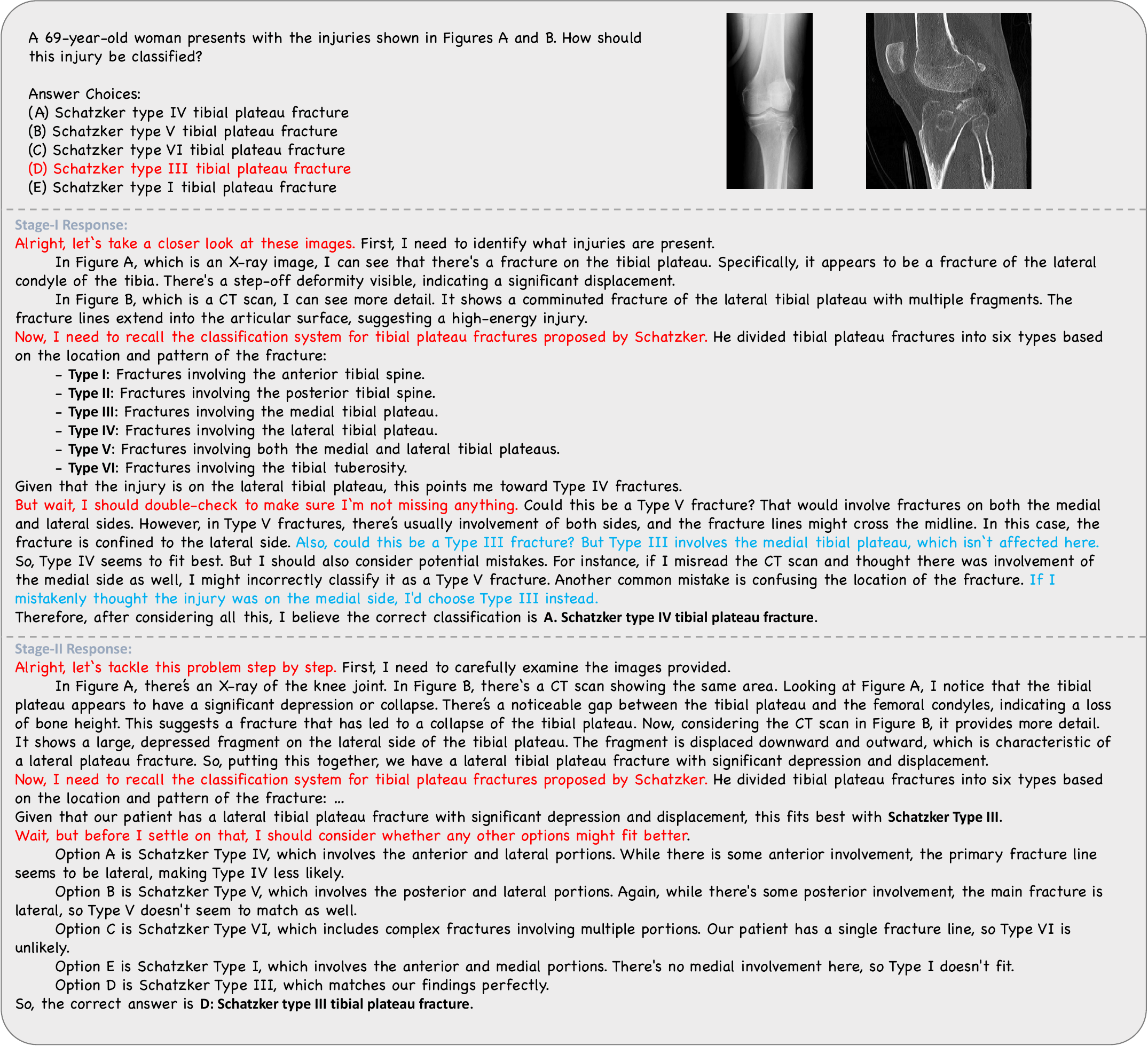}
}
\vspace{-10pt}
\caption{Case illustration of the differences in the responses of models in Stage-I and Stage-II. \textcolor{cyan!50}{Blue text} in the figure indicates visual hallucinations generated by the models.}
\label{fig:example_app3}
\end{figure}

\end{document}